\documentclass{article}
\usepackage{spconf,amsmath,graphicx}
\usepackage{tabularx}
\usepackage[square,sort,comma,numbers]{natbib}
\usepackage{color}
\usepackage{hyperref}

\title{Real-Time Automatic Fetal Brain Extraction in Fetal MRI by Deep Learning}
\name{Seyed Sadegh Mohseni Salehi$^{\star \dagger}$, Seyed Raein Hashemi$^{\star \ddagger}$, Clemente Velasco-Annis$^{\star}$,\sthanks{This study was supported in part by NIH grants R01 EB018988, R01 EB013248, and R03 DE022109; the McKnight Foundation, and the Brain and Behavior Research Foundation. The trained model associated with this work is available at \href{https://bitbucket.org/bchradiology/u-net}{\color{blue} bitbucket.org/bchradiology/u-net}.}}
\secondlinename{Abdelhakim Ouaalam$^{\star}$, Judy A. Estroff$^{\star}$, Deniz Erdogmus$^{\dagger}$, Simon K. Warfield$^{\star}$, Ali Gholipour$^{\star}$}
\address{$^{\star}$ Computational Radiology Laboratory, Boston Children's Hospital, and Harvard Medical School\\$^{\dagger}$ Electrical and Computer Engineering Department, Northeastern University, Boston, MA, USA\\$^{\ddagger}$ Computer and Information Science Department, Northeastern University, Boston, MA, USA}
\begin{document}
%
\maketitle{}
\begin{abstract}
Brain segmentation is a fundamental first step in neuroimage analysis. In the case of fetal MRI, it is particularly challenging and important due to the arbitrary orientation of the fetus, organs that surround the fetal head, and intermittent fetal motion. Several promising methods have been proposed but are limited in their performance in challenging cases and in real-time segmentation. We aimed to develop a fully automatic segmentation method that independently segments sections of the fetal brain in 2D fetal MRI slices in real-time. To this end, we developed and evaluated a deep fully convolutional neural network based on 2D U-net and autocontext, and compared it to two alternative fast methods based on 1) a voxelwise fully convolutional network and 2) a method based on SIFT features, random forest and conditional random field. We trained the networks with manual brain masks on 250 stacks of training images, and tested on 17 stacks of normal fetal brain images as well as 18 stacks of extremely challenging cases based on extreme motion, noise, and severely abnormal brain shape. Experimental results show that our U-net approach outperformed the other methods and achieved average Dice metrics of $96.52\%$ and $78.83\%$ in the normal and challenging test sets, respectively. With an unprecedented performance and a test run time of about 1 second, our network can be used to segment the fetal brain in real-time while fetal MRI slices are being acquired. This can enable real-time motion tracking, motion detection, and 3D reconstruction of fetal brain MRI.  
\end{abstract}
\begin{keywords}
U-net, fetal brain segmentation, fetal MRI, brain extraction, convolutional neural network
\end{keywords}
\section{Introduction}
\label{sec:intro}

Fetal magnetic resonance imaging (MRI) is clinically performed to evaluate the fetus in cases with suspicious or detected abnormalities on prenatal ultrasound~\cite{glenn2010mr}. Fetal MRI has also been instrumental in monitoring and characterizing fetal brain development. Nonetheless, MRI is very susceptible to motion and fetuses move significantly during MRI scans. To minimize the effects of motion, fetal MRI is performed through fast 2D snapshot imaging; i.e. fast 2D \textit{slices} are acquired one after each other to form a \textit{stack} of slices. Each slice acquisition, which takes between 1-2 seconds provides a good-quality image of a section of the anatomy, but the stack of slices does not represent coherent 3D anatomic boundaries due to inter-slice motion artifacts~\cite{gholipour2014fetal}.

\indent Recent advances in fetal MRI processing have shown that 3D images of the fetal brain can be reconstructed through inter-slice motion correction and super-resolution volume reconstruction~\cite{gholipour2010robust,kuklisova2012reconstruction,kainz2015fast,tourbier2017automated}. These techniques assume a rigid model for fetal head motion thus require manually or automatically detected brain masks. Accurate brain extraction has shown to improve inter-slice motion correction and volume reconstruction~\cite{tourbier2017automated,keraudren2014automated}. The need for brain masks has motivated a series of studies into fetal brain extraction in fetal MRI.

Keraudren et al. used bundled SIFT (bSIFT) features to detect the brain in fetal MRI~\cite{keraudren2013localisation} and segmented the brain using an approach based on random forests and conditional random field (RF-CRF)~\cite{keraudren2014automated}. Taimouri et al.~\cite{taimouri2015template} proposed a block matching approach to simultaneously detect a bounding box around the brain and match the orientation of the brain to a template. Due to a search in the space of possible orientations, this technique was computationally very expensive. Tourbier et al.~\cite{tourbier2015automatic} proposed an atlas-based fetal brain segmentation approach that required a predefined bounding box around the brain. This approach was also very time consuming as it relied on deformable registration to multiple atlases. For fast fully automatic brain segmentation, Khalili et al.~\cite{khalili2017automatic} recently proposed a voxelwise convolutional neural network adopted from~\cite{moeskops2016automatic}. They augmented their neural network output with post-processing based on the connected components algorithm to generate brain masks. All of the above methods reported average Dice metrics slightly above $90\%$.

To further improve segmentation accuracy, extend its use, and achieve real-time performance by avoiding any post-processing in this work, we have adopted a deep fully convolutional neural network based on the U-net~\cite{ronneberger2015u}, which has recently shown great performance in 3D brain image segmentation~\cite{salehi2017auto}. While the technique in~\cite{salehi2017auto} focused on segmenting the brain in 3D reconstructed images, our goal here is to segment brain sections independently on original 2D fetal MRI slices, which is much more challenging due to motion and 2D appearance of brain sections. To this end, we trained our network using manually-drawn brain masks on 250 fetal MRI stacks from 26 subjects, and compared its performance on typical and extremely challenging test sets against a 3-pathway voxelwise network plus post processing as suggested in~\cite{khalili2017automatic}, as well as the bSIFT-RF-CRF method based on~\cite{keraudren2014automated}. Experimental results show that our proposed method based on U-net outperforms the other methods and achieves a test run time of less than 1 second, which makes it particularly attractive and useful for real-time application.

\section{Methods}
\label{sec:methods}

\subsection{A 2D U-net architecture}
\label{ssec:subsubhead}

We designed and evaluated our first fully convolutional network~\cite{long2015fully, shelhamer2017fully} based on the U-net architecture~\cite{ronneberger2015u} and Auto-Net~\cite{salehi2017auto}. We describe the details of this network here.

The U-net style architecture consists of a contracting path and an expanding path. The contracting path contains padded $3\times3$ convolutions followed by ReLU non-linear layers. A $2\times2$ max pooling operation with stride 2 is applied after every two convolutional layers. After each downsampling by the max pooling layers, the number of features is doubled. In the expanding path, a $2\times2$ upsampling operation is applied after every two convolutional layers, and the resulting feature map is concatenated to the corresponding feature map from the contracting path. At the final layer a $1\times1$ convolution with linear output is used to reach the feature map with a depth equal to the number of classes (brain or non-brain tissue).

The output layer in the network consists of 2 planes (one per class). We applied softmax along each pixel to form the loss. We did this by reshaping the output into a  $ width\times height\times 2$ matrix and then applying cross entropy. To balance the training samples between classes we calculated the total cost by computing the weighted mean of each class. The weights are inversely proportional to the probability of each class appearance, i.e. higher appearance probabilities led to lower weights. Cost minimization on 100 epochs was performed using ADAM optimizer~\cite{kingma2014adam} with an initial learning rate of 0.0001 multiplied by 0.9 every 2000 steps. The training time for this network was approximately three hours on a workstation with an Nvidia Geforce GTX1080 GPU.

We did not use 3D U-net~\cite{cciccek20163d,salehi2017tversky} here because 1) inter-slice motion and motion-induced artifacts lead to structural boundaries that are incoherent in 3D, 2) slices are occasionally disrupted by fast motion and appear dark, and 3) slice thickness is usually much larger than the in-plane resolution. We aimed to segment brain sections on slices independently to detect good-quality brain structures and motion-corrupted slices, therefore we chose a 2D U-net architecture rather than its 3D counterpart.


\subsection{A voxelwise approach}
\label{ssec:voxelwise}

We also designed and evaluated a voxelwise network architecture~\cite{salehi2017auto} adopting a 3-pathway method and post processing based on the connected components algorithm both suggested in~\cite{khalili2017automatic}, as well as a morphological closing operation as the final step. To accelerate testing, we used fully convolutional layers instead of the fully connected layers used in ~\cite{khalili2017automatic}. We describe the details of this network here.

The voxelwise architecture consists of three fully convolutional paths. Each of these pathways contains four convolutional layers followed by a ReLU nonlinear function and batch normalization. In the first layer, 24 kernels of size $5 \times 5 $ were used for the patch sizes of $15 \times 15$ and $25 \times 25$, and $7 \times 7$ kernels for the patches of size $51 \times 51$. The second layer contained 32 kernels of sizes $3 \times 3$, $3 \times 3$ and $5 \times 5$ on patches respectively. The third layer was designed to have 48 kernel sizes of $3 \times 3$ for all patches. In the proposed architecture, fully convolutional layers were used instead of fully connected layers~\cite{sermanet2013overfeat} to achieve much faster testing time, as the whole image can be tested in a network with convolutional layers while voxels are tested in a network with fully
connected layers~\cite{salehi2017auto}. This fully convolutional method does not use any pre-processing method like brain localization, and is much faster in training and testing compared to fully connected methods such as~\cite{khalili2017automatic}. We also argue that doing a post processing like the 3D connected components is not necessarily a more efficient and accurate way than using pre-processing procedures like localization.

We used 30,000 samples per image and for each sample used three 2D patches with different sizes of $51 \times 51, 25 \times 25$ and $15 \times 15$ to collect local and global context information. These patch sizes were shown to perform best in a 2.5D network for 3D brain image segmentation~\cite{salehi2017auto}. The training process took 2 hours on an Nvidia Geforce GTX 1080 GPU, passing 3 epochs using ADAM optimizer~\cite{kingma2014adam} with an initial learning rate of 0.00005. Early stopping was used to prevent overfitting and decrease the time of unnecessary training.


\section{Results}
\label{sec:results}

Data for this study was obtained from fetal MRI scans performed on 3T Siemens Skyra scanners at Boston Children's Hospital, where the Institutional Review Board committee approved the imaging protocols. For each subject multiple single shot fast spin echo (SSFSE) images were acquired with in-plane resolution of 1 to 1.25 mm, and slice thickness of 2 to 4 mm. The gestational age of the fetuses at the time of scans was between 22 to 38 weeks (mean=29, stdev=5). In total, 285 stacks were selected from 33 fetal MRI sessions. The fidelity of data selection was reviewed by a radiologist with specialty expertise in fetal MRI. A trained research assistant precisely and manually segmented the fetal brain in each slice of all stacks. These were considered ground truth segmentations for training and testing. A total of 250 stacks from 26 sessions were selected as training data, 17 stacks from 2 typical sessions were selected as normal test data, and 18 stacks from 5 very challenging sessions were selected as challenging test data. The challenging sessions involved cases with extreme motion, a case with very abnormal brain shape due to dural sinus fistula, and a case with extremely noisy images.

\begin{figure}
    \centering
    \includegraphics[width=0.85\columnwidth]{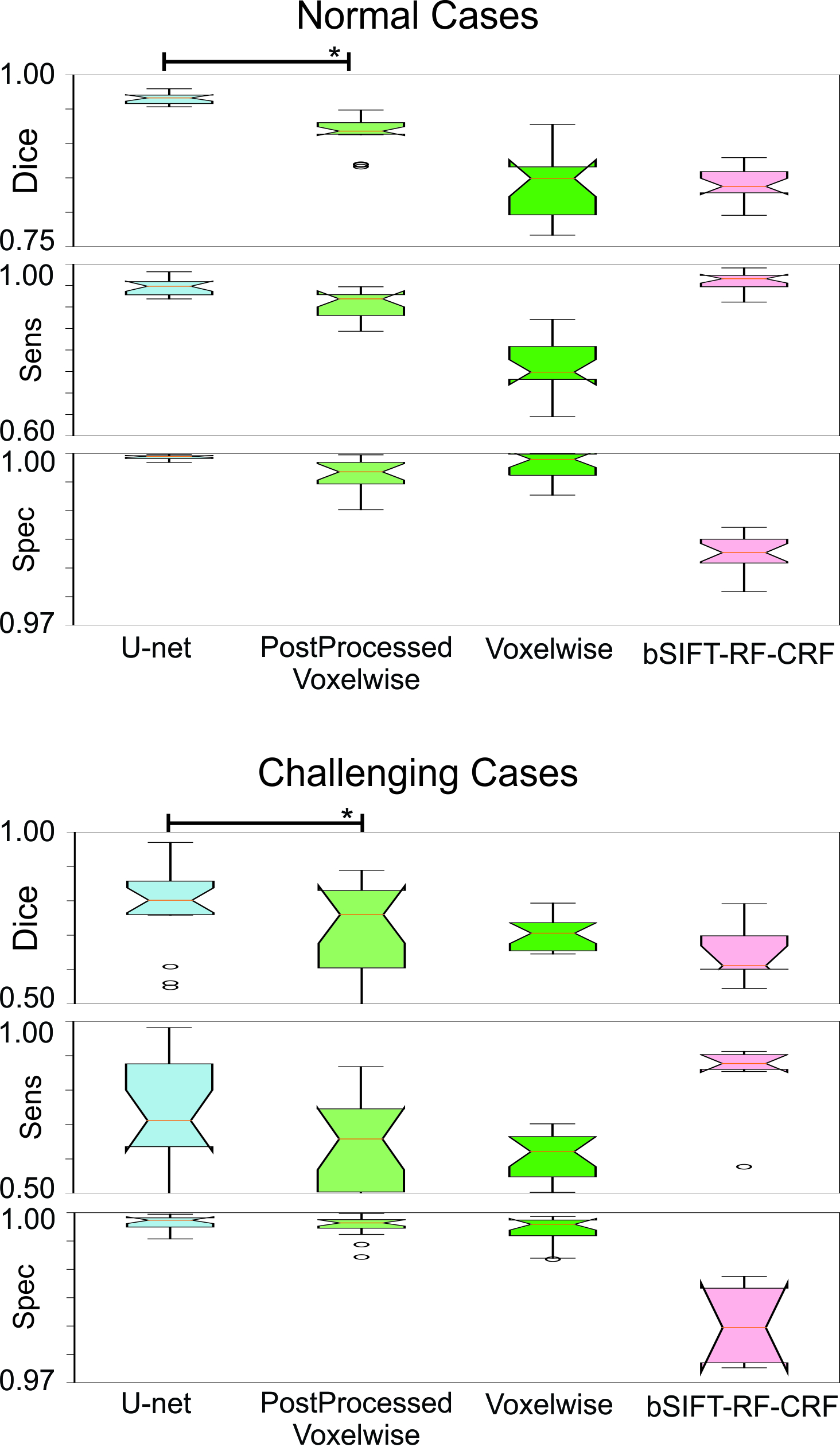}
    \caption{Boxplots of evaluation scores (Dice, sensitivity, specificity) for 2 test sets (normal and challenging cases). The deep learning methods worked for all stacks for all cases, but the bSIFT-RF-CRF method failed in 11 out of 18 stacks (3 out of 5 challenging cases) so the challenging test results for bSIFT-RF-CRF are based on 7 stacks from 2 cases only. The U-net based method shows the best scores.}
    \label{fig:plotbox}
\end{figure}

\begin{figure*}[!ht]
    \centering
    \includegraphics[width=0.95\textwidth]{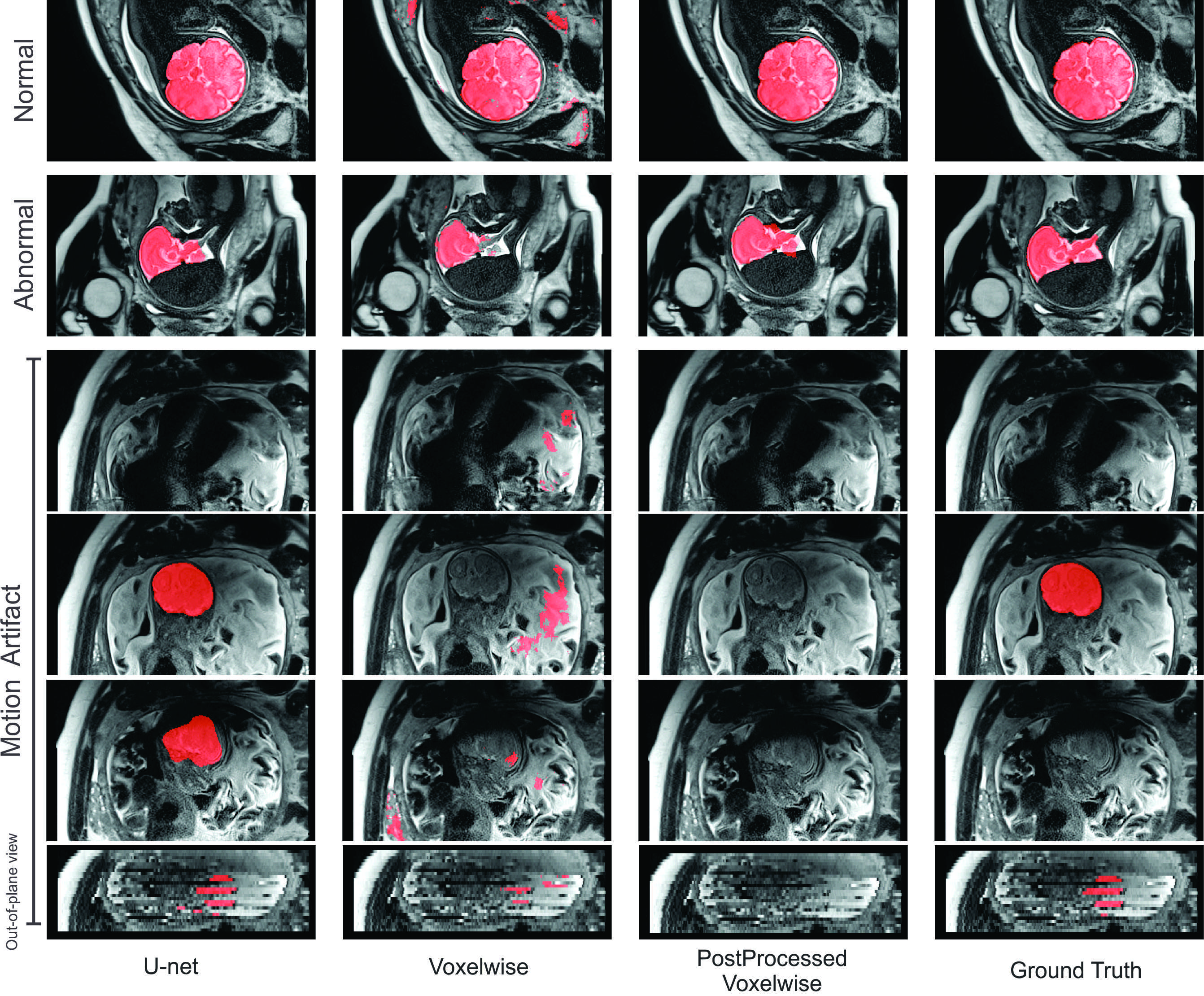}
    \caption{Predicted brain masks overlaid on original fetal MRI for a normal case (top row), an abnormal case (second row), and a case with extreme motion (last four rows showing 3 consecutive slices and an out-of-plane view of the same stack); using U-net (first column), the voxelwise method (second column), voxelwise after post processing (third column), and the manually drawn ground truth (last column). The U-net approach performed much better than the voxelwise method. Note that U-net detected part of the brain in a motion-corrupted scan which was not segmented by the expert due to perceived motion corruption.}
    \label{fig:example}
\end{figure*}

To evaluate and compare performance of 3 fast methods, Dice overlap coefficient was used to compare the predicted brain mask $P$ with the ground truth mask $R$, as $ D = \frac{2\left | P\cap R \right |}{\left | P \right |+\left | R \right |} = \frac{2TP}{2TP+FP+FN}$,
where $TP$, $FP$, and $FN$ are the true positive, false positive, and false negative rates, respectively. We also evaluate specificity, $\frac{TN}{TN+FP}$, and sensitivity, $\frac{TP}{TP+FN}$, of the network.

Table~\ref{table:one} shows the results of the U-net based method to those of the voxelwise approach (before and after post-processing) and the bSIFT-RF-CRF method proposed in~\cite{keraudren2014automated}. Our U-net based method shows the highest Dice coefficient, with an increase of about 5\% compared to the post processed voxelwise approach. The average testing time for U-net was 0.915 second per stack (less than 50 ms per slice). This is considered real-time since each SSFSE slice takes about 1-2 seconds to acquire on the MRI scanner.  

Figure 1 shows boxplots of Dice, sensitivity and specificity of all the analyzed methods for normal and challenging test datasets. In Figure 2, we demonstrate slices of the results of the U-net and voxelwise methods for three cases: a case from the normal test set; a case with severe brain abnormality; and three adjacent slices of a case with extreme motion as well as an out-of-plane view of these images. Although there are lots of inter-slice motion, the U-net based method was capable of segmenting the brain quite accurately. The upsampling layers and concatenations in U-net helped effectively learning multi-scale information on 2D slices. Overall, both quantitative and qualitative evaluation results indicate that U-net performed significantly better than the 3-pathway voxelwise method in brain segmentation. We note that the bSIFT-RF-CRF method failed in 11 out of 18 challenging test stacks in the brain detection step presumably due to the failure of a RANSAC process to reject the influence of outliers. These cases with severely abnormal brain shape and extreme motion pose challenges for such statistically-driven methods.

The original voxelwise Dice score on the normal test set was $84.37\% \pm 0.05$. After applying the connected components algorithm we achieved a Dice score of $85.29\% \pm 0.03$; and finally after a morphological closing operation with a radius of 5 voxels, we achieved average Dice score of $91.66\% \pm 0.02$, sensitivity of $90.64\% \pm 0.03$ and specificity of $99.62\% \pm 0.002$ as shown in Table~\ref{table:one}. The average testing time for each slice was about 1 second, and the average time for post processing was about 15 seconds per stack.



\begin{table}
\begin{center}
    \vspace{-0.2cm}
    \caption{Comparing Dice, sensitivity and specificity for normal test set, showing 5\% higher average Dice score for U-net compared to the voxelwise method after post processing (PP).}
    \begin{tabularx}{.49\textwidth}{lllll}
    \hline
    Method & Dice & Sensitivity & Specificity & Time\\ \hline
    Our 2D U-net & \textbf{96.52\%} & \textbf{94.60\%} & \textbf{99.92\%} & \textbf{0.9s}\\ \hline
    Voxelwise-PP & 91.66\% & 90.64\% & 99.62\% & 45s\\ \hline
    Voxelwise & 84.37\% & 76.30\% & 99.77\% & 30s\\ \hline
    SIFT-RF-CRF & 83.94\% & 74.50\% & 99.80\% & 60s \\ \hline
    \end{tabularx}
    \label{table:one}
    \vspace{-1.0cm}
\end{center}
\end{table}

\section{Conclusion}
\label{sec:conclusion}
We developed and tested a 2D U-net and a voxelwise fully convolutional network to automatically segment fetal brain in fetal MRI. For the voxelwise approach, we used post processing methods that significantly boosted our prediction accuracy on normal cases; but the U-net approach achieved much better results on both normal and challenging test sets with a testing time of about 1 second per stack without any post-processing. The accuracy of our fast methods in segmenting the intracranial region of fetal MRI can lead to improved methods of motion detection and correction, brain segmentation, and reconstruction which are subjects of ongoing work.

\begin{table}
\begin{center}
    \vspace{-0.2cm}
    \caption{Comparing the evaluation scores for the challenging test set, showing 12\% higher average Dice score for U-net compared to the voxelwise method with post-processing.}
    \begin{tabularx}{.49\textwidth}{lllr}
    \hline
    Method & Dice & Sensitivity & Specificity\\ \hline
    U-net & \textbf{78.83\%} & 71.97\% & \textbf{99.82\%} \\ \hline
    Voxelwise-PP & 66.87\% & 58.66\% & 99.76\% \\ \hline
    Voxelwise & 66.32\% & 57.21\% & 99.68\% \\ \hline
    SIFT-RF-CRF* & 64.95\% & \textbf{84.19}\% & 98.02\% \\ \hline
    \end{tabularx}
    \label{table:two}
    \vspace{-0.5cm}
\end{center}
\end{table}




\section{References}
\label{sec:refer}
\begingroup
    \renewcommand{\section}[2]{}%
    \setlength{\bibsep}{1pt}
    \bibliographystyle{IEEEbib}
    \bibliography{refs}
\endgroup

\end{document}